# Visual Appearance Analysis of Forest Scenes for Monocular SLAM

James Garforth[1] and Barbara Webb[1]

*Abstract*— Monocular simultaneous localisation and mapping (SLAM) is a cheap and energy efficient way to enable Unmanned Aerial Vehicles (UAVs) to safely navigate managed forests and gather data crucial for monitoring tree health. SLAM research, however, has mostly been conducted in structured human environments, and as such is poorly adapted to unstructured forests. In this paper, we compare the performance of state of the art monocular SLAM systems on forest data and use visual appearance statistics to characterise the differences between forests and other environments, including a photorealistic simulated forest. We find that SLAM systems struggle with all but the most straightforward forest terrain and identify key attributes (lighting changes and in-scene motion) which distinguish forest scenes from "classic" urban datasets. These differences offer an insight into what makes forests harder to map and open the way for targeted improvements. We also demonstrate that even simulations that look impressive to the human eye can fail to properly reflect the difficult attributes of the environment they simulate, and provide suggestions for more closely mimicking natural scenes.

## I. INTRODUCTION

This paper looks at the problem of performing visual Simultaneous Localisation and Mapping (SLAM) in unstructured natural environments (such as forests), rather than the structured man-made environments (such as offices and city streets) that play host to the majority of SLAM research [1], [2], [3]. Our target application is forestry, where data gathering to assess tree health could be greatly enhanced in efficiency, scale and accuracy if robots could navigate within forests. Unmanned aerial vehicles (UAVs) have the agility to traverse uneven or vegetation-cluttered terrain and inspect trees at any height, but in this scenario might not be able to carry many sensors, so our focus is on monocular SLAM. There are a number of reasons to expect that SLAM may be difficult in forests. Global Positioning System (GPS) data tends to be unreliable under canopy cover. Vegetation is locally dynamic due to wind and patchy light, yet has high global similarity that could lead to substantial aliasing. The ability of state of the art SLAM algorithms to deal with such conditions is largely untested. Our main contributions are:

- Qualitative analysis of the performance of monocular SLAM algorithms in forest environments.
- Characterisation of fundamental visual differences between forests and more traditionally mapped environments (e.g. offices) using scene statistics.
- Assessment of photorealistic simulation as a groundtruthing environment for developing SLAM algorithms.

*This work was supported by the Edinburgh Centre for Robotics and the Engineering and Physical Sciences Research Council.
[1]James Garforth and Barbara Webb are with the School of Informatics, University of Edinburgh, 10 Crichton Street, Edinburgh, EH8 9AB, United Kingdom. {james.garforth,b.webb}@ed.ac.uk

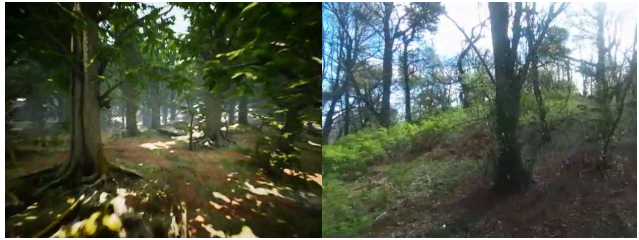

Fig. 1: Our photorealistic simulated forest (Left) and a region of real forest from our Hillwood dataset (Right).

## II. BACKGROUND

### A. Mapping Forests

Forest management is greatly aided by accurate measurement of trees, but below the canopy, few survey methods have been successfully automated. Humans with measuring tapes are still the standard method for collecting tree trunk diameter data, for example. Scanning lasers have been used to gather higher quality data but still require humans to walk the area to be mapped, carrying heavy equipment instead of a tape. This approach also requires substantial post-processing of point clouds. In [4] points are clustered into individual trees; aligned to an existing map, fine tuned by ICP, and diameters at a fixed height extracted. Pierzchala et al.[5] improve the performance of offline laser maps in forests with loop closure techniques from SLAM.

Miettinen et al.[6] developed a real-time laser-based system, mounted on a mobile platform, but found it very difficult to match individual trees in all but the sparsest managed forests. Ohman et al.[7] added cameras to this system, as the visual appearance of bark on tree trunks makes them easier to differentiate. They use a phase congruency edge detector to deal better with low contrast images, but were still unable to obtain usable levels of accuracy and processing speed.

The depth camera of Google's Tango device has been shown to provide sufficiently accurate reconstruction of forest scenes[8], but only on a small scale.

### B. Monocular SLAM

A key advance in the field of monocular SLAM was Klein and Murray's Parallel Tracking and Mapping (PTAM) [9], which has been successfully used on UAVs [10], [11]. PTAM, as with most monocular SLAM systems after it, is a "keyframe" based method. Instead of storing sensor data from every single camera frame, PTAM selects a smaller set of frames that it judges most representative, and builds the map from these. Mapping only needs to occur when keyframes are added, so can be run in a separate thread from camera motion tracking, improving real-time performance.

"ORB-SLAM"[1] adds faster feature comparison and greater view invariance to PTAM by using ORB features[12], extracted once and used for the tracking, mapping and loop closure subsystems. Instead of extracting features, "direct" SLAM methods, e.g. "LSD-SLAM"[13], use a metric to sub-select useful pixels and perform alignment between images by minimizing photometric error. The volume of pixels used varies from selecting all of them ("Dense" methods like DTAM[14]) through to sparsely selected pixels in methods like "SVO"[15] and "DSO"[16]. Yang et al.[17] note that direct methods are more robust than feature-based when scenes have low texture and provide a more complete reconstruction due to their use of more image information, but are vulnerable to camera properties and rapid lighting changes.

Recently, ORB-SLAM has been used on a Bebop UAV [18] for monitoring crops in plantations, combined with GPS and Inertial Measurement Unit (IMU) data for navigation. Smolyanskiy et al.[19] use DSO for obstacle detection on a UAV performing autonomous trail following in forests. Though neither addresses the full complexity of forest survey, they indicate that ORB-SLAM and DSO are good starting points for experimentation.

### C. Adapting SLAM to Environment Properties

The application of monocular SLAM in forests may require adaptation of the algorithms to the characteristics of the visual environment. Previous researchers have looked at methods for basic classification of visual environments to support switching between SLAM methods [20] [21]. In more directly relevant work, Saeedi et al.[22] perform "Design Space Exploration", searching through the parameter space of a SLAM algorithm for a desired trade-off between accuracy and efficiency, given a high level description of the environment. They use Kullback-Leibler (KL) divergence, an information theoretic measure of how different two distributions (in this case, intensity histograms) are from one another, and posit that this one statistic neatly encompasses the variation of structure and motion within the scene. Research into the adaptation of animal vision to natural environments [23], [24] suggests additional relevant 'scene statistics' might be luminance, contrast and colour distributions. In particular, lighting-related effects are well known to impact SLAM[25], with potential solutions explored in [26] and [27].

## III. METHODS

### A. Methodological Issues

We wish to evaluate the performance of mapping algorithms in a forest environment, but several issues make this difficult to do rigorously. Evaluation of the accuracy of pose estimation requires a ground truth pose, usually obtained from accurate external sensors, such as a calibrated camera rig or, outdoors, GPS. Forest canopy cover, however, interferes with GPS signals. Placement of some other form of tracking sensors at ground level in a large area of interest is costly, and if visible would inadvertently provide beacons that influence the performance of the systems we want to test. Evaluating 3D reconstruction also requires ground truth, usually obtained either by working in an environment of known, rigid structure, or by using a more complex but comprehensive mapping solution than the one being assessed (for example a high powered laser scanner). However, mapping of large natural environments with such scanning systems remains a difficult problem[28] and as such does not guarantee an error free ground truth.

An alternative method for achieving pose and reconstruction ground truths is to perform experiments in a simulated environment where both are known perfectly. Simulation would also provide much greater control over key factors expected to affect performance, such as lighting and wind, allowing for comparison of different conditions. The trade-off is the possibility that a simulated forest does not properly capture the traits that make real forests challenging, and that improvements developed in simulation will not transfer to real world applications.

### B. Datasets

We put together a selection of video datasets (summarised in Table I), including forest environments, structured urban environments and simulations, which would allow us to assess mapping algorithms and compare visual properties. The datasets come in a number of formats, such as video files, folders of individual frames or in the "rosbag" format used by the Robot Operating System[29].

The first of our forest data is from the SFU Mountain Dataset[30] (henceforth simply SFU), recorded from a wheeled robot driving a forested mountain road in a variety of weather and lighting conditions. We use only the "Dry" conditions to avoid our results also reflecting the effects of weather. Partway through each video, the vehicle moves from open road to a canopy-covered dirt track. As we are primarily interested in how the latter scenario differs from the former, we split the video at this point, forming "SFU Road" and "SFU Forest". We use the left of the vehicle's two forward facing cameras as a single video stream.

We also recorded our own forest videos, referred to here as Hillwood. This dataset contains videos recorded from two low cost UAV platforms (Parrot's AR.Drone and Bebop) and these are analysed separately. Hillwood provides more complex camera motion than SFU, as the camera takes winding routes and retraces the same area. It also contains sequences away from any path, over and around vegetation, to test performance of SLAM in the absence of any clear man-made structure such as a track.

We chose two "classic" datasets to represent the more structured environments typically used for testing SLAM applications: the TUM Monocular dataset[31], recorded on a hand held camera; and KITTI[32], recorded from a car on city streets. We further sample from TUM Monocular to make two datasets: one indoor (offices) and one outdoor (urban). For purpose of comparison, we also include Bebop Indoor, an office video recorded on the same platform as Hillwood Bebop.

Additionally, we recorded a simulated dataset utilising the Unreal Engine and a set of photorealistic forest assets created for film and video game rendering[33], manually flying the virtual camera through it. Although in principle this should resemble the Hillwood data (a drone travelling through unstructured forest) we note that in practice this produced much smoother motion than a real UAV.

## C. SLAM System Comparison

We selected four state of the art monocular SLAM systems for comparison: (1) ORBSLAM2[1], a sparse feature-based method, (2) LSD-SLAM[13] and (3) DSO[16], which are both direct methods and (4) SVO[15], which is semi-direct, using direct methods for tracking and then using features later in its pipeline. We intended to also include OKVIS [34], but our chosen datasets do not provide the required inertial measurements or calibration data. We intended to assess the benefits of visual inertial SLAM in future.

The selected systems were all calibrated for and run on the Hillwood and SFU datasets to assess their performance on forest data. Demonstration of these systems on indoor and outdoor scenes can be found in their original papers.

## D. Visual Appearance Metrics

Forest environments could have some global visual properties that differ from structured environments and thus set a challenge for visual SLAM. To assess this, we use these statistics to compare the datasets:

1) **Lighting changes** as we expect that the effect of the forest canopy will be frequent large switches between bright sunlight and shadow. We measure both luminance and contrast. Luminance is defined as the average intensity value of a frame. We use Root Mean Squared Contrast, defined as the standard deviation of luminance in a frame, divided by the mean. Change is recorded as the difference between each subsequent pair of frames for each of these statistics.

2) **Kullback-Leibler divergence** of intensity histograms, as an approximation of scene structure and motion[22]. We expect that the heavily deformable nature of forests (leaves, grasses etc.) leads to a large amount of motion in the scene with even a small amount of wind. KL Divergence for intensity images is calculated as in [22].

$$D_{KL}(I_t \parallel I_{t-1}) = \sum_{t=0}^{256} I_t(u) \log \frac{I_t(u)}{I_{t-1}(u)}$$

Where $I_t$ and $I_{t-1}$ indicate the normalised intensity histograms (256 bins) of the frames.

3) **Variance of the Laplacian** approximates the frequency and strength of edges within the image. In this way it can give us information about two traits of our datasets: firstly, how in focus the images are (as excessive camera motion will cause blurring) and secondly how complex the textures in the scene are. Implemented using OpenCV's Laplacian function, we also rescale all images to the same size (320x240) beforehand as image size effects the result.

We also looked at two "secondary" statistics, which help to exclude non-environment specific attributes of datasets from being the major factors in our results:

1) **Features Matches** are used as a simplified measure of the ease of tracking features for visual SLAM in the absence of ground truth data for our datasets. SIFT features are extracted and matches sought between subsequent pairs of frames (100 features per frame). We use a brute force matcher, then a ratio test to decide which matches to accept.

2) **Reprojected Similarity** estimates frame to frame overlap in order rule out the possibility that large differences in the primary statistics are due to large camera motions. The similarity is calculated as the Mean Squared Error of two subsequent frames after using feature matches to reproject them into the same frame of reference.

These measures were implemented in Python using OpenCV. We calculate all of the change statistics for subsequent pairs of frames, sampling the datasets at the same frame rate (10fps) to account for the expectation that a camera sampling faster will see smaller changes between frames. Our pipeline converts images to intensity (grey) and normalises before it calculates the statistics. It is also worth noting that we skip the first 30 frames of each video because in some datasets these contain artefacts from the camera's auto-calibration which can skew the results.

## IV. RESULTS

### A. SLAM System Performance

We tested four SLAM methods on the forest datasets: Hillwood AR, Hillwood Bebop and SFU Forest. The only fully tracked real sequences were obtained using either ORBSLAM2 or DSO on the SFU "dry" video (see Figure 2b). When run on Hillwood data, ORBSLAM2 and DSO achieved tracking for a small portion of the video (less than 1 minute), resulting in very small sections of map that we could not confirm the quality of by eye. LSD-SLAM and SVO fail to start tracking on any of the forest videos, either producing no map or one in which features are distributed with no recognisable structure.

For both ORBSLAM2 and DSO, the tracked pose drifts in scale over time, as is most notable from the misalignment of the outbound and inbound tracks. This is not unexpected from monocular SLAM systems, but the fact that neither system manages to recognise previously visited locations and correct the drift demonstrates a failure of their respective place recognition mechanisms. We note that the demonstrated tracking from these two systems shows there is no specific advantage in this environment for feature-based (ORBSLAM2) vs. direct (DSO) methods. Looking at the point clouds produced by the two systems in Figure 2c, however, we can see that the larger volume of data afforded by DSO's direct method does provide a more detailed reconstruction.

The failure of two of our four evaluated systems to even initialise pose tracking on any of the tested datasets demonstrates the difficulty of forest environments for even short term mapping. The two systems that were able to maintain tracking only did so on the SFU Forest video, and only if we substantially increased the number of features extracted per image (on ORBSLAM2 we use 3000 vs. the

TABLE I: A summary of the datasets reviewed, with video count, frame rate, resolution and a representative example frame.

| Name | Hillwood AR | Hillwood Bebop | SFU Forest | SFU Road | Bebop Indoor | TUM Indoor | TUM Outdoor | KITTI | Unreal |
|---|---|---|---|---|---|---|---|---|---|
| Count | 3 | 1 | 1 | 1 | 1 | 10 | 6 | 10 | 5 |
| FPS | 15 | 30 | 30 | 30 | 30 | 50 | 30 | 10 | 30 |
| Res | 320x240 | 1920x1080 | 752x480 | 752x480 | 1920x1080 | 1280x1024 | 1280x1024 | 1241x376 | 640x480 |
| Eg. | | | | | | | | | |

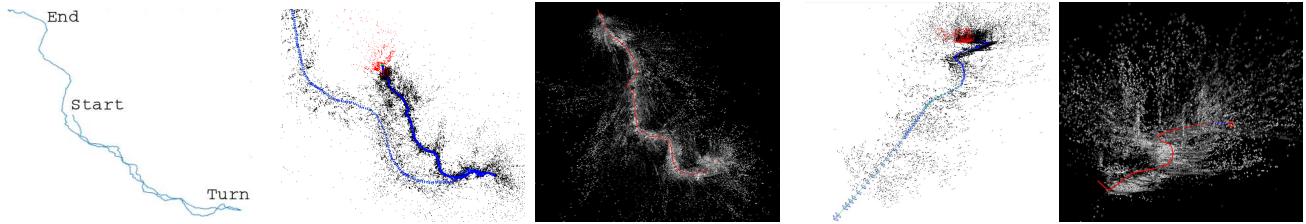

(a) GPS data provides rough ground truth but is inaccurate under canopy.

(b) Top-down views exhibiting the recovered route. Both track the outbound route well, but suffer scale drift on the return route, which for ORBSLAM2 appears shorter and for DSO stacks frames together.

(c) Horizontal views exhibiting the level of reconstruction. DSO's denser map recovers notably more structure.

Fig. 2: The full tracks and point clouds as produced by ORBSLAM2 (white background) and DSO (black background) on the forest video where they are most successful (SFU forest under dry conditions).

default 1000), suggesting that in this environment a large number of candidate features need to be extracted at each frame in order to ensure sufficient crossover between frames for matching.

The successful SFU video represents an easy use case for monocular SLAM, as the camera is mounted to a slowly moving ground vehicle and experiences very little roll, pitch or even yaw and all viewpoint changes happen gradually. The vehicle also sticks to a clear, well-defined dirt path, and observation of the point cloud produced by ORBSLAM2 especially (Figure 2b) seems to suggest that the ground here is providing the majority of the tracked features. We note that ORBSLAM2 and DSO also did not appear to find tracking difficult on Unreal data, but as we discuss in the next section these videos are perhaps failing to reflect the difficulties of forest scenes, so they were not compared extensively.

In Hillwood videos, where all algorithms struggle, the camera was mounted on a UAV which follows a less straight-forward route with more rotational motion in all axes. But tracking failures occur even on relatively straight sections, suggesting camera motion is not the only factor causing problems. Notably, these videos contain a lot less clear ground than SFU, instead often travelling over vegetation or fallen leaves. The UAV also frequently flies near to and between trees, leading to regular occlusions of parts of the scene, while the ground robot in SFU usually keeps enough distance from trees that this effect would be greatly reduced.

### B. Visual Appearance Comparison

In the previous section we found that all the SLAM systems tested failed on the more challenging forest data, yet they have previously been established as effective systems in more typical scenarios such as indoor mapping and city roads. Hence, we did an initial investigation of whether there are any general differences in the scene statistics of these different scenarios (see methods).

*1) Lighting Changes:* Luminance (Figure 3a) and contrast (Figure 3b) changes do appear to differentiate between the forest datasets (Hillwood, SFU Forest) and most of the classic ones (TUM). The median differences for luminance and contrast are higher in the forest videos, irrespective of the platform they were recorded from, and the distribution is also larger, reflecting a tendency of these datasets towards both generally larger lighting changes over time and larger sudden lighting changes. KITTI is the only non forest dataset to see a similar distribution of lighting changes, but it is also the only dataset where the camera is travelling faster than walking speed. This large speed difference could account for the large visual appearance changes between frames.

When comparing Hillwood Bebop and Bebop Indoor the camera parameters are identical but the results differ, each following the general trend for the other datasets in their respective environment. Likewise, when we compare SFU Forest and SFU Road, where possible complicating factors such as time of day or camera motion and parameters are ruled out, we still see major lighting differences, supporting the possibility that the forest canopy is responsible for the lighting variability.

The Unreal dataset demonstrates the lowest median and the tightest distribution of values for luminance and contrast changes, likely caused by the simulator's lighting model. Even though the simulation is lit by directional lighting from an artificial sun, it also has ambient lighting. As a result, the areas of the simulated forest that are in shadow still appear relatively well lit, meaning the camera is less likely to experience large swings from light to dark as it transitions between direct sunlight and shadow.

*2) Kullback-Leibler Divergence:* It is clear from the KL divergence results (Figure 3c) that the datasets in forested areas (including SFU Road, which is lined with trees) are less predictable frame to frame than classic environment datasets are. The difference between the two SFU datasets indicates that going off road and under canopy with the same platform markedly reduces the predictability further. All of this goes to support the idea that vegetation has a notable impact on how much scenes change over time, perhaps due to the amount of small-scale motion (e.g. of leaves) they introduce. The higher KLD values for the Bebop Hillwood data suggest that this is not the only factor, however. The other videos mostly travel forwards along a path, while the Bebop goes back and forth over one area, so it is very likely that the large amount of rotational motion by the Bebop also contributes to the KL divergence being considerably higher.

The very low KL divergence seen in the Unreal simulated data implies a high predictability of each frame given the previous one. It is not obvious to the human eye that this environment contains any less motion or complex structure than the real forest, in a way that would account for less predictability. We suspect that the difference may be in the camera model. The simulated camera does not suffer from any noise and applies considerable blurring, both of which serve to improve the similarity between subsequent images.

*3) Variance of the Laplacian:* The results for the variance of the Laplacian correlate well with the observed complexity in the datasets, such that the texture-heavy natural scenes sit at one end of the spectrum and indoor scenes (with flat textureless regions like walls) are at the other. A potential limitation of this measure is demonstrated by SFU Road, however, as the prominence of a large featureless sky throughout this video is the cause of its low median variance (confirmed by rerunning the pipeline with the top half of the video cropped). It is interesting to note that the high texture datasets are the ones that are more difficult for visual SLAM, as this typically benefits from textured scenes for feature extraction. It is likely that this visual complexity overwhelms feature extraction and matching and would explain why getting ORBSLAM2 to work required us to increase the number of candidate features extracted.

There is little evidence in these results to suggest that blur (which would reduce the variance) is a problem in any of the real world data, as the datasets at most risk should have been those with a less stable camera (such as Hillwood, which reports the highest variances). The low variance for the simulated Unreal data, is likely caused by the game engine adding too much motion blur, anti-aliasing, or using a limited colour palette.

*4) Feature Matches:* The distinction between forest and classic datasets is less clear with respect to the percentage of good frame to frame feature matches, but there does still appear to be some trend. Both Hillwood datasets and SFU Forest achieve fewer median successful matches between frames, as well as having lower minimum matches. These results would imply that it is indeed harder to extract and continue to track features reliably from forest scenes than urban ones.

Similarly to the results with lighting changes, it makes sense that KITTI's lower results here would be caused by

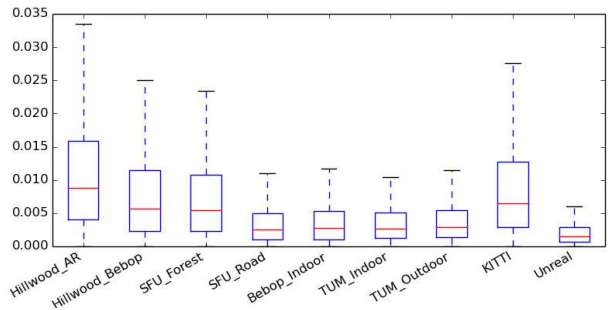

(a) Luminance changes between subsequent frames. Higher median and maximum changes for the three datasets under forest canopy on the left, but lower for the simulated Unreal forest.

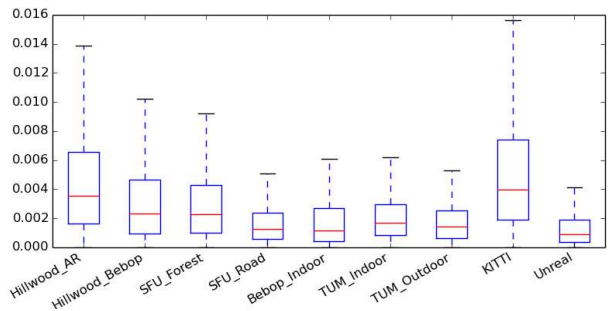

(b) Contrast changes between subsequent frames. Higher median and maximum changes for the three datasets under forest canopy on the left, but lower for the simulated Unreal forest.

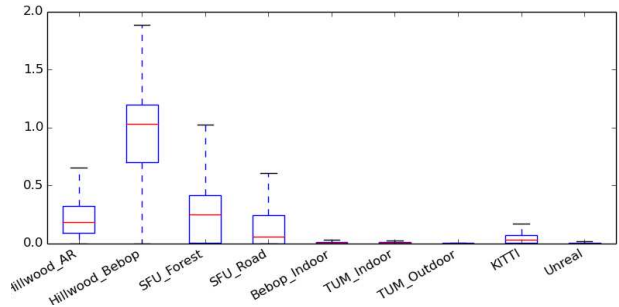

(c) Kullback-Leibler divergence between subsequent frames. Higher median and maximum changes for the four vegetation heavy datasets on the left, but not for the simulated Unreal forest.

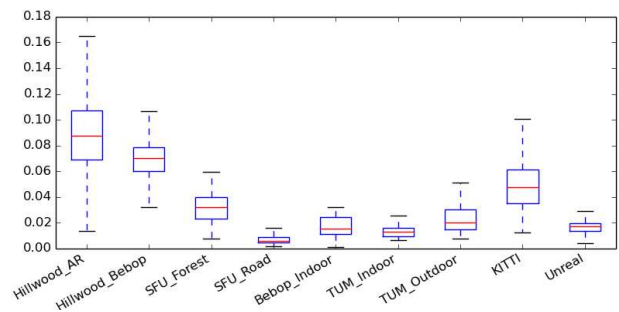

(d) Variance of the Laplacian for all frames. The higher variance for outdoor (especially forested) datasets suggests the presence of more and stronger edges/texture.

Fig. 3: Our primary statistics characterise differences between video datasets gathered in forest and urban environments.

the camera moving further between frames.

SFU Road achieves consistently high matches. As the skyline is always visible in this data, and has been shown to be useful for navigation[35], it is expected that this is providing a large number of reliable features. To test this, we reran the pipeline with the top half of each image removed, and found that the results for SFU Road did indeed become more like the other datasets.

Notably, this statistic is also the only one of those tested where the Unreal data does not stand out significantly from the real forest scenes.

*5) Reprojected Similarity:* After using matched features to reproject subsequent pairs of images into the same frame of reference, we see very little difference between most of the datasets. Seeing similar levels of overlap between frames in these datasets helps reject the idea that the other results reported here (primarily for KLD) could be caused by significant rotational or translational motion specific to the forest datasets, rather than an attribute of the environment. KITTI displays a much higher error after reprojection, as expected for a dataset where the camera is moving significantly faster and overlap would be expected to be smaller. SFU Road and our simulation, however, have notably lower errors than other datasets despite not being notably slower. The explanation is likely similar to IV-B.3 in that the sky, or the game engine's limited colour palette, result in self-similarity across the environment.

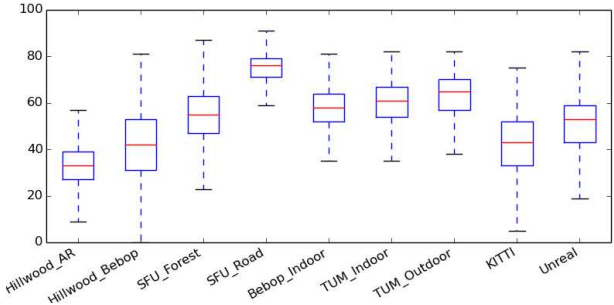

(a) Frame to frame SIFT feature matches, after ratio test, out of 100. Less reliable matching for forest datasets supports our observations of poor SLAM performance.

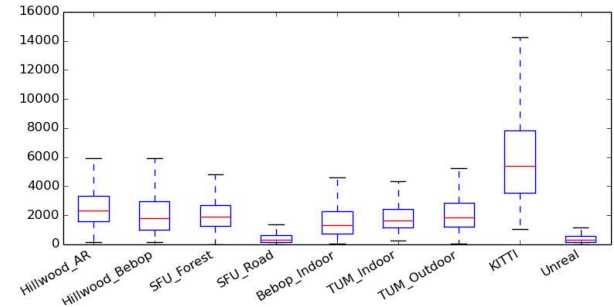

(b) Frame to frame overlap, as measured by Mean Squared Error after reprojection of each frame into the next. There is no notable difference between the overlap of frames in forest and non forest datasets.

Fig. 4: Secondary statistics, used as support for other claims rather than to directly characterise environments.

## V. CONCLUSION AND FUTURE WORK

In this paper we evaluated the performance of state of the art monocular SLAM in forests. Such unstructured natural environments have not been traditionally considered in SLAM evaluations, despite the high potential for robot applications in this domain. We found that even with tuning of the parameters, only two systems (ORBSLAM2 and DSO) successfully ran, and only on the easiest of our test cases: a slow and steady ground robot travelling along a clear path in a forest. In most cases, systems failed to produce a usable map. These tests identify the unsuitability of any existing solutions for off the shelf use in this domain. One particular problem area observed is loop closure, which we hope to address with place recognition Convolutional Neural Networks in future work.

Improving the performance of existing systems requires an understanding of how the forest environment differs from the standard use-cases. We performed statistical analyses of forest and non-forest data and found some key differences. Lighting (represented by luminance and contrast) changes over time distinguish forests from offices and roads and are likely caused by the gaps and movement of the canopy leading to frequent variation in the amount of sunlight illuminating the scene. In-scene motion (represented by entropy measure KL divergence) is also notably higher in forest scenes, likely due to the presence of wind and flexible vegetation. This suggests that the key developments needed (to extend ORBSLAM or DSO which worked in the simpler case, or in new SLAM algorithms) are methods to deal more robustly with lighting and scene dynamics. These are already active areas of interest but gain additional motivation from our analysis.

We also investigated the use of highly realistic game engine based simulation as an alternative to real world data when testing an improved SLAM system for natural environments. We note that such a simulation would be useful in a number of ways, providing ground truth data that is hard to match in the real world and also allowing fine tuned control over the exact variables (light and motion) that we want to control for. We issue a warning, however, against abandoning real world data too quickly, as our scene statistics mark out the simulated forest as more different in appearance from real forests than urban environments are.

Finally, rigorous evaluation of SLAM systems for forests requires more complete test data than simply video. If a solution can be found to ground-truthing in real forests, perhaps by careful synchronisation with lidar data, then this can be used to create a new forest dataset. Alternatively, improvements can be made to our existing simulation, for example through the addition of realistic sensor models, and our statistical approach can be used to establish if a greater resemblance to real forest data has been achieved.

## ACKNOWLEDGMENT

This work was undertaken with advice and guidance from forest mapping specialists Carbomap (carbomap.com).

## REFERENCES

[1] R. Mur-Artal, J. Montiel, and J. Tardos, "ORB-SLAM: a Versatile and Accurate Monocular SLAM System," *arXiv preprint arXiv:1502.00956*, 2015.

[2] M. Milford and G. F. Wyeth, "SeqSLAM: Visual route-based navigation for sunny summer days and stormy winter nights," *Proceedings - IEEE International Conference on Robotics and Automation*, pp. 1643–1649, 2012.

[3] R. Newcombe, A. J. Davison, S. Izadi, P. Kohli, O. Hilliges, J. Shotton, D. Molyneaux, S. Hodges, D. Kim, and A. Fitzgibbon, "KinectFusion: Real-time dense surface mapping and tracking," *2011 10th IEEE International Symposium on Mixed and Augmented Reality*, pp. 127–136, 2011.

[4] T. Takashi, A. Asuka, M. Toshihiko, K. Shuhei, S. Keiko, M. Mitsuhiro, T. Shuhei, N. Shuichi, M. Akiko, C. Yukihiro, S. Kouji, and H. Toru, "Forest 3D Mapping and Tree Sizes Measurement for Forest Management Based on Sensing Technology for Mobile Robots," *Springer Tracts in Advanced Robotics*, vol. 92, pp. 357–368, 2014.

[5] M. Pierzchała, P. Giguère, and R. Astrup, "Mapping forests using an unmanned ground vehicle with 3d lidar and graph-slam," *Computers and Electronics in Agriculture*, vol. 145, pp. 217–225, 2018.

[6] M. Miettinen, M. Ohman, A. Visala, and P. Forsman, "Simultaneous Localization and Mapping for Forest Harvesters," *Proceedings 2007 IEEE International Conference on Robotics and Automation*, no. April, pp. 517–522, 2007.

[7] M. Ohman and M. M. Kosti Kannas, Jaakko Jutila, Arto Visala and Pekka Forsman, "Tree Measurement and Simultaneous Localization and Mapping System for Forest Harvester," *Field and Service Robotics Springer Tracts in Advanced Robotics*, vol. 42, pp. 369–378, 2008.

[8] J. Tomaštík, Š. Saloň, D. Tunák, F. Chudỳ, and M. Kardoš, "Tango in forests–an initial experience of the use of the new google technology in connection with forest inventory tasks," *Computers and Electronics in Agriculture*, vol. 141, pp. 109–117, 2017.

[9] G. Klein and D. Murray, "Parallel Tracking and Mapping for Small AR Workspaces," in *2007 6th IEEE and ACM International Symposium on Mixed and Augmented Reality*, IEEE. IEEE, nov 2007, pp. 1–10.

[10] S. Weiss, M. W. Achtelik, S. Lynen, M. C. Achtelik, L. Kneip, M. Chli, and R. Siegwart, "Monocular Vision for Long-term Micro Aerial Vehicle State Estimation: A Compendium," *Journal of Field Robotics*, vol. 30, no. 5, pp. 803–831, sep 2013.

[11] D.-N. Ta, K. Ok, and F. Dellaert, "Monocular Parallel Tracking and Mapping with Odometry Fusion for MAV Navigation in Feature-lacking Environments," *Intelligent Robots and Systems ( . . . )*, 2013.

[12] E. Rublee, V. Rabaud, K. Konolige, and G. Bradski, "ORB: An efficient alternative to SIFT or SURF," *Proceedings of the IEEE International Conference on Computer Vision*, pp. 2564–2571, 2011.

[13] J. Engel, T. Schöps, and D. Cremers, "LSD-SLAM: Large-Scale Direct Monocular SLAM," in *Computer Vision ECCV 2014*, ser. Lecture Notes in Computer Science, D. Fleet, T. Pajdla, B. Schiele, and T. Tuytelaars, Eds. Cham: Springer International Publishing, 2014, vol. 8690, pp. 834–849.

[14] R. Newcombe, S. J. Lovegrove, and A. J. Davison, "DTAM: Dense tracking and mapping in real-time," in *2011 International Conference on Computer Vision*, IEEE. IEEE, nov 2011, pp. 2320–2327.

[15] C. Forster, M. Pizzoli, and D. Scaramuzza, "Svo: Fast semi-direct monocular visual odometry," in *Robotics and Automation (ICRA), 2014 IEEE International Conference on*. IEEE, 2014, pp. 15–22.

[16] J. Engel, V. Koltun, and D. Cremers, "Direct sparse odometry," *IEEE transactions on pattern analysis and machine intelligence*, 2017.

[17] N. Yang, R. Wang, and D. Cremers, "Feature-based or Direct: An Evaluation of Monocular Visual Odometry," pp. 1–12, 2017.

[18] V. Duggal, M. Sukhwani, K. Bipin, G. S. Reddy, and K. M. Krishna, "Plantation Monitoring and Yield Estimation using Autonomous Quadcopter for Precision Agriculture," 2016.

[19] N. Smolyanskiy, A. Kamenev, J. Smith, and S. Birchfield, "Toward Low-Flying Autonomous MAV Trail Navigation using Deep Neural Networks for Environmental Awareness," 2017.

[20] J. Collier and A. Ramirez-Serrano, "Environment classification for indoor/outdoor robotic mapping," *Proceedings of the 2009 Canadian Conference on Computer and Robot Vision, CRV 2009*, pp. 276–283, 2009.

[21] D. C. Asmar, J. S. Zelek, and S. M. Abdallah, "SmartSLAM : localization and mapping across multi-environments," *Systems, Man and Cybernetics, 2004 IEEE International Conference on*, pp. 5240 – 5245, 2004.

[22] S. Saeedi, L. Nardi, E. Johns, B. Bodin, P. H. J. Kelly, and A. Davison, "Application-oriented Design Space Exploration for SLAM Algorithms," pp. 1–8.

[23] E. P. Simoncelli and B. A. Olshausen, "Natural Image Statistics and Neural Representation," *Annual Review Neuroscience*, 2001.

[24] W. S. Geisler, "Visual Perception and the Statistical Properties of Natural Scenes," 2008.

[25] C. Valgren and A. J. Lilienthal, "Sift, surf and seasons: Long-term outdoor localization using local features," in *3rd European conference on mobile robots, ECMR'07, September 19-21, Freiburg, Germany*, 2007, pp. 253–258.

[26] N. Yang, R. Wang, X. Gao, and D. Cremers, "Challenges in monocular visual odometry: Photometric calibration, motion bias, and rolling shutter effect," *IEEE Robotics and Automation Letters*, vol. 3, no. 4, pp. 2878–2885, 2018.

[27] P. Kim, B. Coltin, O. Alexandrov, and H. J. Kim, "Robust visual localization in changing lighting conditions," in *Robotics and Automation (ICRA), 2017 IEEE International Conference on*. IEEE, 2017, pp. 5447–5452.

[28] B. Risse, M. Mangan, W. Stürzl, and B. Webb, "Software to convert terrestrial lidar scans of natural environments into photorealistic meshes," *Environmental Modelling & Software*, vol. 99, pp. 88–100, 2018.

[29] M. Quigley, K. Conley, B. Gerkey, J. Faust, T. Foote, J. Leibs, R. Wheeler, and A. Y. Ng, "Ros: an open-source robot operating system," in *ICRA workshop on open source software*, vol. 3, no. 3.2. Kobe, Japan, 2009, p. 5.

[30] J. Bruce, J. Wawerla, and R. Vaughan, "The SFU mountain dataset: Semi-structured woodland trails under changing environmental conditions," in *IEEE Int. Conf. on Robotics and Automation 2015, Workshop on Visual Place Recognition in Changing Environments*, 2015.

[31] J. Engel, V. Usenko, and D. Cremers, "A photometrically calibrated benchmark for monocular visual odometry," in *arXiv:1607.02555*, July 2016.

[32] A. Geiger, P. Lenz, and R. Urtasun, "Are we ready for autonomous driving? the kitti vision benchmark suite," in *Conference on Computer Vision and Pattern Recognition (CVPR)*, 2012.

[33] M. U. GmbH. Broadleaf forest collection. [Online]. Available: https://www.youtube.com/watch?v=Zyq_UpOQ9r4

[34] S. Leutenegger, S. Lynen, M. Bosse, R. Siegwart, and P. Furgale, "Keyframe-based visual-inertial odometry using nonlinear optimization," *International Journal of Robotics Research*, vol. 34, no. 3, pp. 314–334, 2015.

[35] T. Stone, M. Mangan, P. Ardin, B. Webb *et al.*, "Sky segmentation with ultraviolet images can be used for navigation," in *Robotics: Science and Systems*. Robotics: Science and Systems, 2014.